\pdfoutput=1
\documentclass[a4paper]{llncs}

\usepackage{amssymb}
\usepackage{graphicx}
\usepackage{subfig}
\usepackage{multirow}
\usepackage{array}
\usepackage{authblk}
\usepackage{booktabs}
\newcolumntype{C}[1]{>{\centering\arraybackslash\hspace{0pt}}p{#1}}
\usepackage{url}
\urldef{\mailsa}\path|{alfred.hofmann, ursula.barth, ingrid.haas, frank.holzwarth,|
\urldef{\mailsb}\path|anna.kramer, leonie.kunz, christine.reiss, nicole.sator,|
\urldef{\mailsc}\path|erika.siebert-cole, peter.strasser, lncs}@springer.com|    
\newcommand{\keywords}[1]{\par\addvspace\baselineskip
\noindent\keywordname\enspace\ignorespaces#1}
\urldef{\mails}\path|{woalsdnd, khwan.jung}@vuno.co, sangjunpark@snu.ac.kr| 

\begin{document}

\mainmatter  

\title{Retinal Vessel Segmentation in Fundoscopic Images with Generative Adversarial Networks}


%
%
\author{Jaemin Son$^1$ \and Sang Jun Park$^2$ \and Kyu-Hwan Jung$^1$\footnote{Corresponding author}}

\institute{$^1$Vuno Inc., Seoul, Korea,\\
\path|{woalsdnd, khwan.jung}@vuno.co|\\
$^2$Department of Ophthalmology, Seoul National University College of Medicine,\\ Seoul National University Bundang Hospital, Seongnam, Korea\\
\path|sangjunpark@snu.ac.kr|}
%
%

\maketitle

\begin{abstract}
Retinal vessel segmentation is an indispensable step for automatic detection of retinal diseases with fundoscopic images. Though many approaches have been proposed, existing methods tend to miss fine vessels or allow false positives at terminal branches. Let alone under-segmentation, over-segmentation is also problematic when quantitative studies need to measure the precise width of vessels. In this paper, we present a method that generates the precise map of retinal vessels using generative adversarial training. Our methods achieve dice coefficient of 0.829 on DRIVE dataset and 0.834 on STARE dataset which is the state-of-the-art performance on both datasets. 

\end{abstract}

\section{Introduction}
Analysis of retinal vessel networks provides rich information about conditions of the eyes and general systemic status. Ophthalmologists can detect early signs of increased systemic vascular burden from hypertension and diabetes mellitus as well as vision threatening retinal vascular diseases such as Retinal Vein Occlusion (RVO) and Retinal Artery Occlusion (RAO) from abnormality in the vascular structures. To aid such analysis, automatic vessel segmentation method, especially from fundus images, has been researched extensively.

In early days, many computer vision algorithms approached to this problem from the perspective of signal processing based on the assumption that the vessels follow particular patterns. Canonical examples are heuristic techniques such as line detection~\cite{ricci2007retinal,nguyen2013effective} and hand-crafted feature extraction~\cite{soares2006retinal,zhang2010retinal}. With the advance of machine learning, however, more improved results were obtained with automatic feature learning. For instance, Becker {\it et al.} extracted features automatically using gradient boosting~\cite{becker2013supervised} and Orlando {\it et al.} proposed to introduce fully connected Conditional Random Field (CRF) whose parameters are trained from data with structured Support Vector Machine (SVM).

In recent years, Convolutional Neural Networks (CNNs) have shown outstanding performance in various computer vision tasks. Several studies have already proven that CNNs achieve improved performance in segmenting retinal vessels and even surpass the ability of human experts in multiple datasets~\cite{maninis2016deep,fu2016deepvessel,xie2015holistically,melinvsvcak2015retinal}. However, the segmented vessels from these methods are rather blurry and suffer from false positives around minuscule and faint branches. This is mainly because the objective function of CNNs used in the existing methods only rely on pixel-wise objective functions that compare gold standard images and model-generated images. This is not desirable since it cannot actively accommodate natural vascular structure that resides in fundus images. 

In fact, the vessel segmentation can be considered as an image translation task where an output segmented vessel map is generated from an input fundoscopic image. If the outputs are constrained to resemble the human expert's annotation, clearer and sharper vessel maps can be obtained. Generative Adversarial Networks (GANs) is a framework that enables to create as realistic outputs as the gold standard~\cite{goodfellow2014generative}. GANs consist of two networks, discriminator and generator. While a discriminator tries to distinguish gold standard images from the outputs generated by the generator, the generator tries to generate as realistic outputs as the discriminator cannot differentiate from the gold standard. 

In this paper, we discuss a new approach to retinal vessel segmentation with adversarial networks. Not only do our methods extract clear and sharp vessels with less false positives compared to existing methods but achieve the state-of-the-art performance in two public datasets, namely, DRIVE and STARE. We show that adversarial training can actually improve the quality of the segmentation by training the generator to extract vessel maps that are indistinguishable from vessel maps annotated by human experts.

\section{Proposed Methods}
\subsection{Network Structure}
\begin{figure}[!th]
  \centering
\includegraphics[scale=0.4]{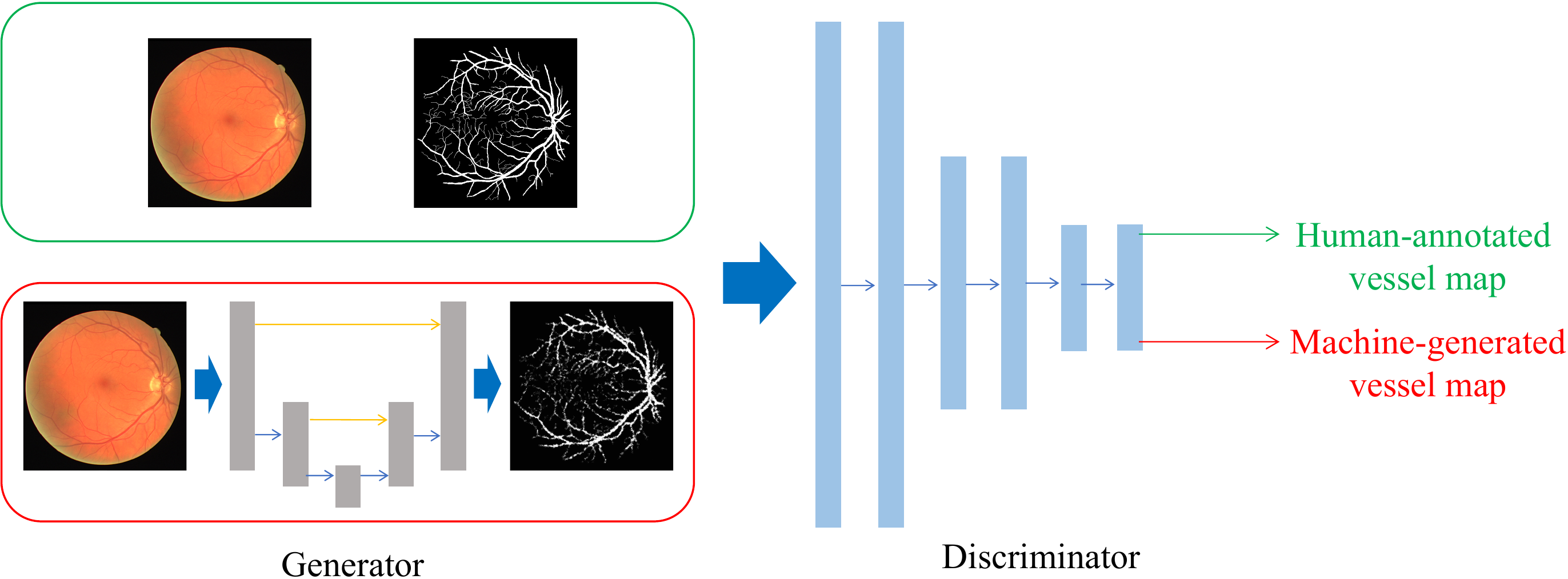}
\caption{The proposed Generative Adversarial Networks (GANs) framework for vessel segmentation. }
\label{fig:GAN}
\end{figure}

In our Generative Adversarial Networks (GANs) setting, the generator is given a fundus image and generates probability maps of retinal vessel with the same size as the input. Values in the probability maps range from 0 to 1 indicating the probability of being a pixel of vessels. The discriminator takes a fundus and vessel image to determine whether the vessel image is the gold standard from the human expert or output of the generator. The overall framework is depicted in Fig.~\ref{fig:GAN}.

For the generator, we follow the spirit of U-Net~\cite{ronneberger2015u} where initial convolutional feature maps are skip-connected~\cite{he2016deep} to upsampled layers from bottleneck layers. This skip-connection is crucial to segmentation tasks as the initial feature maps maintain low-level features such as edges and blobs that can be properly exploited for accurate segmentation. 

As a discriminator, we explored several models with different output size as done in \cite{isola2016image}. In the atomic level, a discriminator determines the authenticity pixel-wise (Pixel GAN) while judgment can also be made in the image level (Image GAN). Between both extremes, it is also possible to set the receptive field to $K\times K$ patch where the decision is made in the patch level (Patch GAN). We investigated Pixel GAN, Image GAN, and Patch GAN with two intermediary patch sizes.

\subsection{Objective Function}
Let the generator $G$ be a mapping from a fundus image $x$ to a vessel image $y$, or $G:x\mapsto y$. Then, the discriminator $D$ maps a pair of $\{x,y\}$ to binary classification $\{0,1\}^N$ where $0$ and $1$ mean that $y$ is machine-generated or human-annotated and $N$ is the number of decisions. Note that $N=1$ for Image GAN and $N=W\times H$ for Pixel GAN with an image of $W\times H$.

Then, the objective function of GAN for the segmentation problem can be formulated as
\begin{equation}
L_{GAN}(G,D)=\mathbb{E}_{x,y\sim p_{data}(x,y)}[\log D(x,y)]+\mathbb{E}_{x\sim p_{data}(x)}[\log (1-D(x,G(x)))].
\end{equation}
Note that $G$ takes input of an image, thus, analogous to conditional GAN~\cite{mirza2014conditional}, but there is no randomness involved in $G$. Then, GAN framework solves the optimization problem of
\begin{equation}
G^*=\arg \min_{G} \big[\max_{D}\mathbb{E}_{x,y\sim p_{data}(x,y)}[\log D(x,y)]+\mathbb{E}_{x\sim p_{data}(x)}[\log (1-D(x,G(x)))]\big].
\end{equation}
For training discriminator $D$ to make correct judgment, $D(x,y)$ needs to be maximized while $D(x,G(x))$ should be minimized. On the other hand, the generator should prevent the discriminator from making correct judgment by producing outputs that are indiscernible to the real data. Since the ultimate goal is to obtain realistic outputs from the generator, the objective function is defined as minimax of the objective.

In fact, the segmentation task can also utilize gold standard images by adding loss functions that penalize distance between the gold standard and outputs such as binary cross entropy
\begin{equation}
L_{SEG}(G)=\mathbb{E}_{x,y\sim p_{data}(x,y)} -y\cdot \log G(x) - (1-y)\cdot\log (1-G(x)).
\end{equation}
Summing up both the GAN objective and the segmentation loss, we can formulate the objective function as
\begin{equation}
G^*=\arg \min_{G} \big[\max_{D} L_{GAN}(G,D)\big] + \lambda L_{SEG}(G)
\label{loss}
\end{equation}
where $\lambda$ balances two objective functions. 

\section{Experiments}
\subsection{Experimental Setup}
Our methods are implemented based on Keras library with tensorflow backend\footnote{Source code is available at \url{https://bitbucket.org/woalsdnd/v-gan}}. We tested our methods on two public datasets, DRIVE and STARE. We trained and tested with the first annotator's vessel images and the compare the performance of the second annotator with our method. For STARE dataset that consists of 20 images, we used the first 10 images for training and tested with the rest as in \cite{maninis2016deep}.

Each image is normalized to z-score for each channel and augmented by left-right flip and rotation. Augmented images are divided into train/validation set with the ratio of 19 to 1 and the models with the least generator loss on the validation set are chosen. We ran multiple rounds of training until convergence in which the discriminator and the generator are trained for an epoch alternatively. We used Adam optimizer with fixed learning rate of $2e{-4}$ and $\beta_1=0.5$ and fix the trade-off coefficient in Eq.\ref{loss} to 10 $(\lambda =10)$.

We evaluate our methods with Area Under Curve for Receiver Operating Characteristic (ROC AUC), Area Under Curve for Precision and Recall Curve (PR AUC) and dice coefficient or F1 measure. For dice coefficient, we thresholded the probability map with Otsu threshold~\cite{otsu1979threshold} that is frequently used in separating foreground and background. For fair measurement, pixels inside the field of view are counted when computing the measures\footnote{In case of STARE dataset that includes no mask images, we generated a mask by detecting a blob in the center.}. 

\subsection{Experimental Results}
Performance of models with different discriminators is compared in Table~\ref{tab:diff_D}. 
\begin{table}[!th]
\centering
\caption{Comparison of models with different discriminators on two datasets with respect to Area Under Curve (AUC) for Receiver Operating Characteristic (ROC), Precision and Recall (PR).}
\begin{tabular}{C{13em} |C{5em} C{5em}| C{5em} C{5em}}
  \toprule
  \multirow{2}{*}{Model}& \multicolumn{2}{c|}{DRIVE}& \multicolumn{2}{c}{STARE}\\ & ROC & PR & ROC& PR \\
  \midrule
  U-Net (No discriminator) &0.9700&0.8867&0.9739& 0.9023\\
  Pixel GAN $(1\times 1)$ &0.9710&0.8892&0.9671&0.8978\\
  Patch GAN-1 $(10\times 10)$&0.9706&0.8898&0.9760&0.9037\\
  Patch GAN-2 $(80\times 80)$&0.9720&0.8933&0.9775&0.9086\\
  Image GAN $(640\times 640)$&{\bf 0.9803}&{\bf 0.9149}&{\bf 0.9838}&{\bf 0.9167}\\
  \bottomrule
\end{tabular}
\label{tab:diff_D}
\end{table}
U-Net, which has no discriminator, shows inferior performance to patch GANs and image GAN suggesting that GANs framework improves quality of segmentation. Also, image GAN, which has the most discriminatory capability, outperforms others. This observation is consistent to claims that a powerful discriminator is key to successful training with GANs~\cite{goodfellow2014generative,radford2015unsupervised}. 

\begin{figure}[!th]
  \centering
\hspace{-3mm}
\subfloat{\includegraphics[scale=0.4]{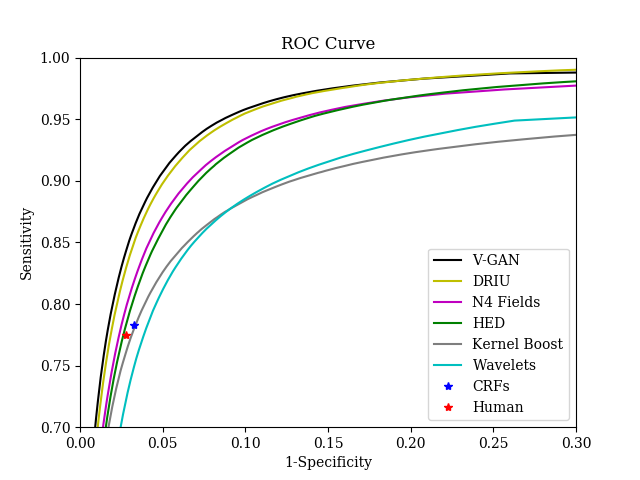}}\hspace{-7mm}
\subfloat{\includegraphics[scale=0.4]{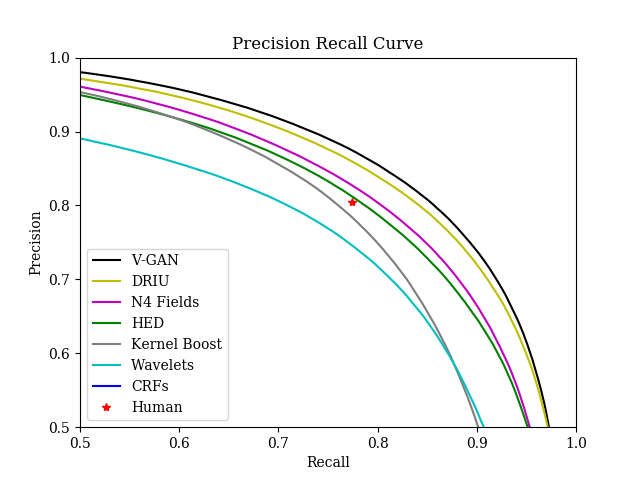}}\\
\vspace{-3mm}
\hspace{-3mm}
\subfloat{\includegraphics[scale=0.4]{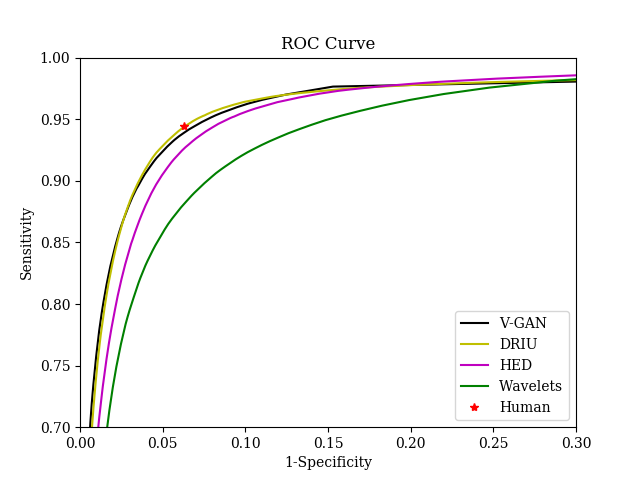}}\hspace{-7mm}
\subfloat{\includegraphics[scale=0.4]{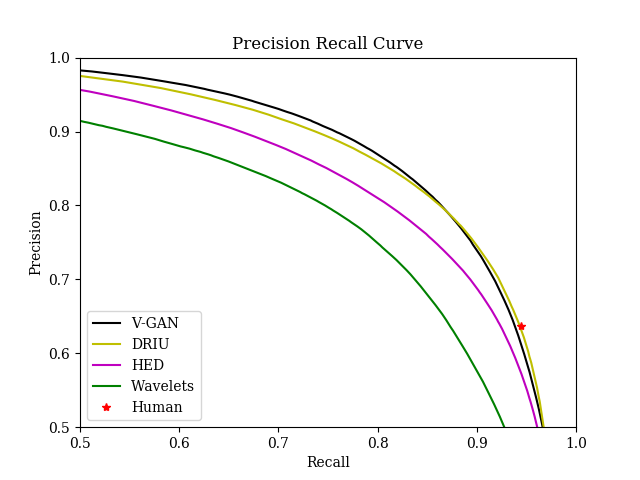}}
\caption{Receiver Operating Characteristic (ROC) curve and Precision and Recall (PR) curve for various methods on DRIVE dataset ({\bf Top}) and STARE dataset ({\bf Bottom}). }
\label{fig:AUC}
\end{figure}

Fig.~\ref{fig:AUC} compares ROC and PR curves for the image GAN (V-GAN) with existing methods and Table~\ref{tab:auc_dice} summarizes AUC for ROC and PR and dice coefficient. We retrieved dice coefficients and output images of other methods from \cite{maninis2016deep} and the curves are computed from the images. Our method shows better performance in other methods in all operating regime except DRIU. Still, our method shows superior AUC and dice coefficient to DRIU. Also, our method surpasses the human annotator's ability on DRIVE dataset.

\begin{table}[!th]
\centering
\caption{Comparison of different methods on two datasets with respect to Area Under Curve (AUC) for Receiver Operating Characteristic (ROC), Precision and Recall (PR) and Dice Coefficient.}
\begin{tabular}{C{9em} |C{4.2em} C{4.2em} C{4.2em} |C{4.2em} C{4.2em} C{4.2em}}
  \toprule
  \multirow{2}{*}{Method}& \multicolumn{3}{c|}{DRIVE}& \multicolumn{3}{c}{STARE}\\ & ROC & PR & Dice&ROC& PR & Dice\\
  \midrule
  Kernel Boost~\cite{becker2013supervised}&0.9306&0.8464&0.800&-&-&-\\
  HED ~\cite{xie2015holistically}&0.9696&0.8773&0.796&0.9764&0.8888&0.805\\
  Wavelets~\cite{soares2006retinal}&0.9436&0.8149&0.762& 0.9694&0.8433&0.774\\
  $N^4$-Fields~\cite{ganin2014n}&0.9686&0.8851&0.805&-&-&-\\
  DRIU~\cite{maninis2016deep}&0.9793&0.9064&0.822&0.9772&0.9101&0.831\\
  Human Expert &-&-&0.791&-&-&0.760 \\
  V-GAN &{\bf 0.9803}&{\bf 0.9149}&{\bf 0.829}&{\bf 0.9838}&{\bf 0.9167}&{\bf 0.834}\\
      \bottomrule
 \end{tabular}
 \label{tab:auc_dice}
\end{table}
 
\begin{figure}[!th]
  \centering
\hspace{-2mm}
\subfloat{\includegraphics[scale=0.145]{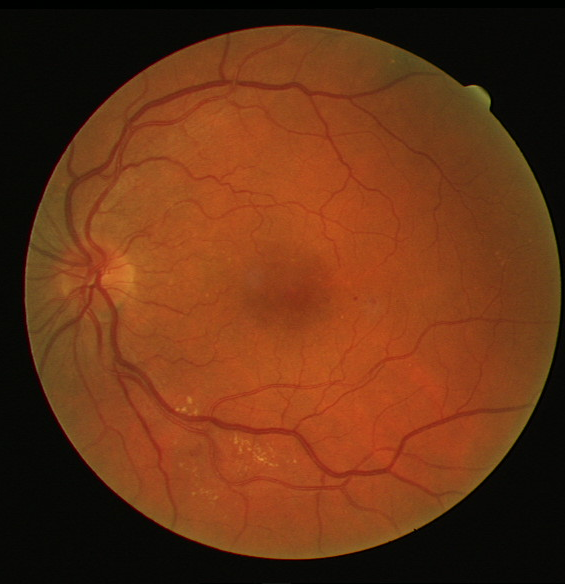}}\hspace{0mm}
\subfloat{\includegraphics[scale=0.145]{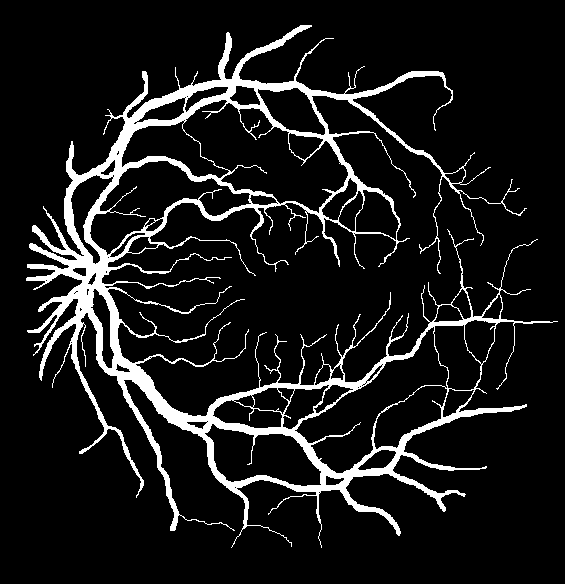}}\hspace{0mm}
\subfloat{\includegraphics[scale=0.145]{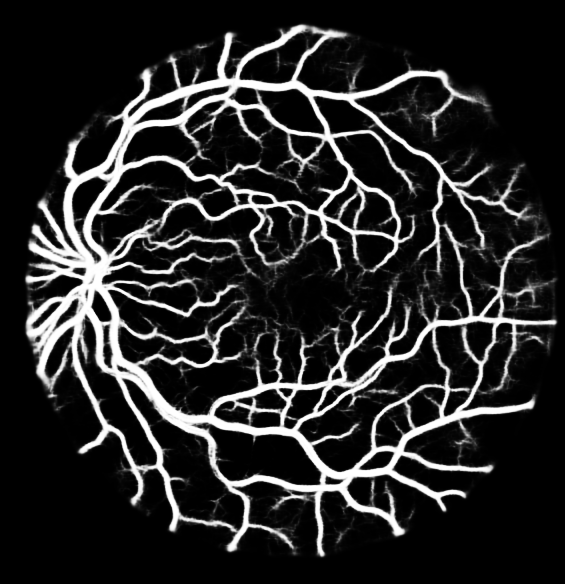}}\hspace{0mm}
\subfloat{\includegraphics[scale=0.145]{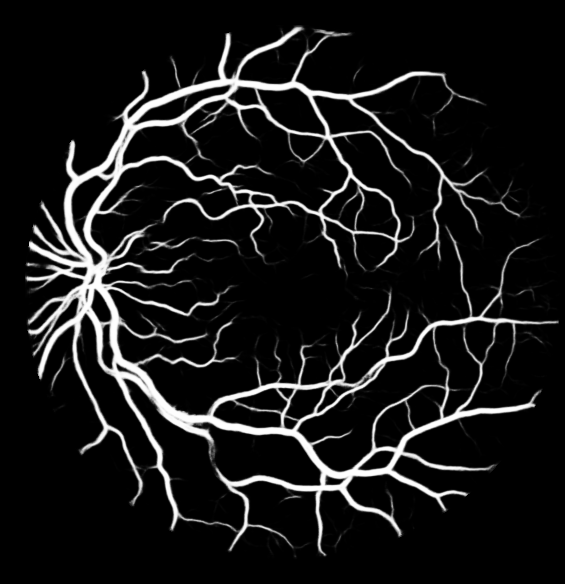}}\hspace{0mm}\\
\hspace{-2mm}
\subfloat{\includegraphics[scale=0.145]{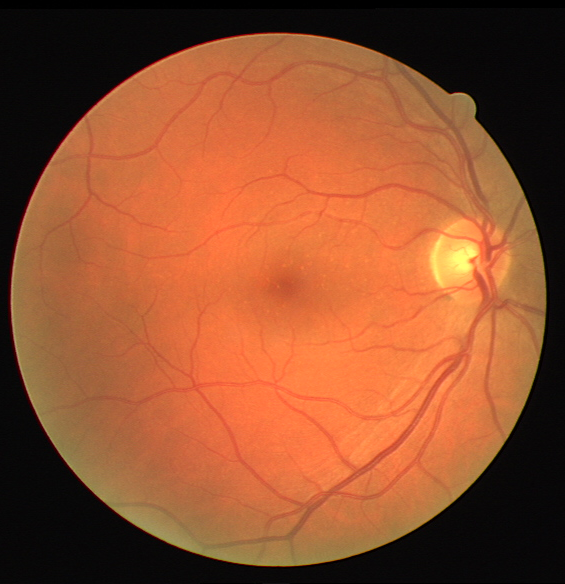}}\hspace{0mm}
\subfloat{\includegraphics[scale=0.145]{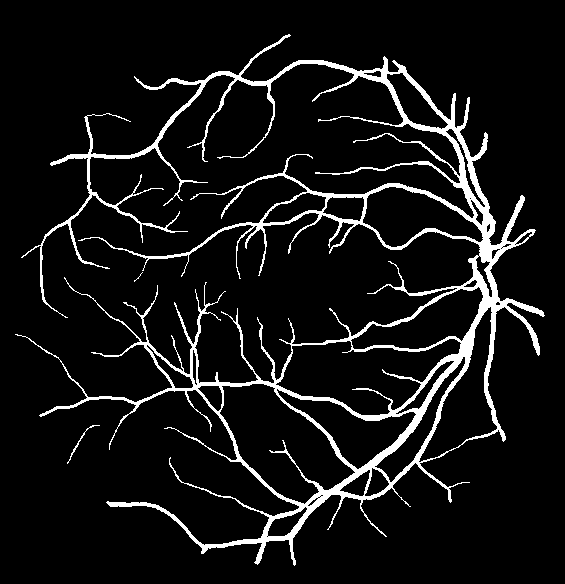}}\hspace{0mm}
\subfloat{\includegraphics[scale=0.145]{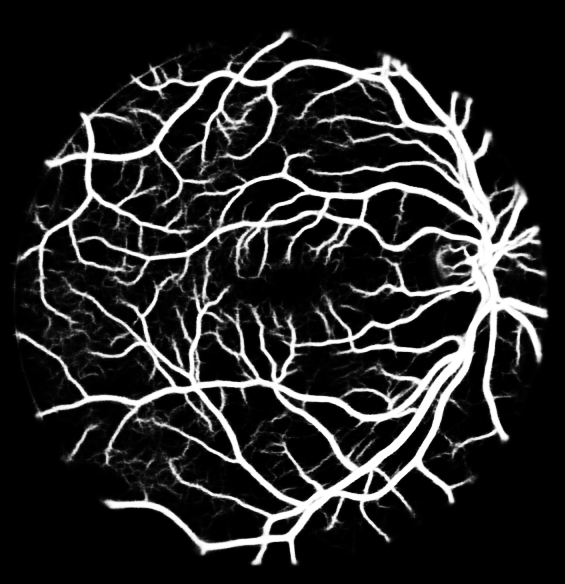}}\hspace{0mm}
\subfloat{\includegraphics[scale=0.145]{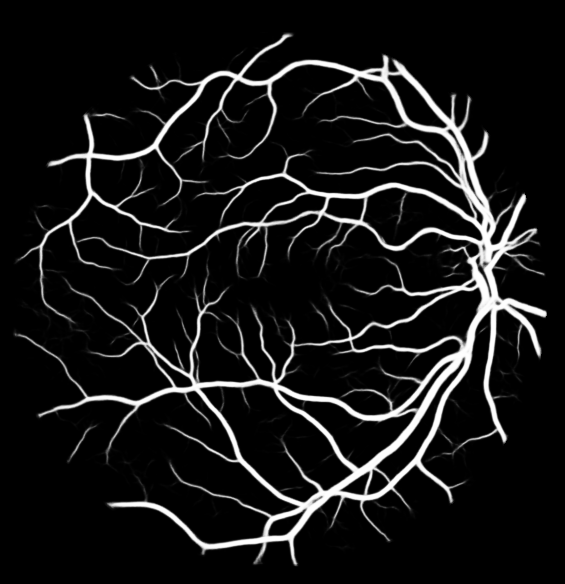}}\hspace{0mm}
\noindent\rule{12cm}{1pt}\\
\vspace{3mm}
\hspace{-2mm}
\subfloat{\includegraphics[scale=0.12]{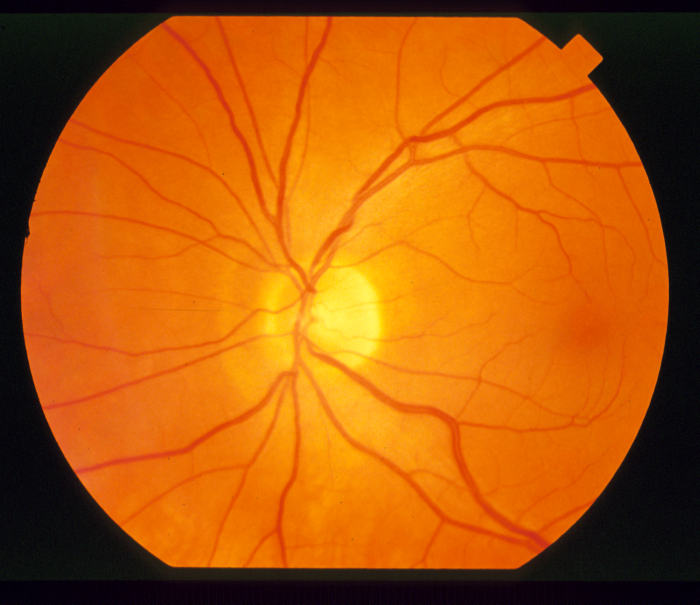}}\hspace{0mm}
\subfloat{\includegraphics[scale=0.12]{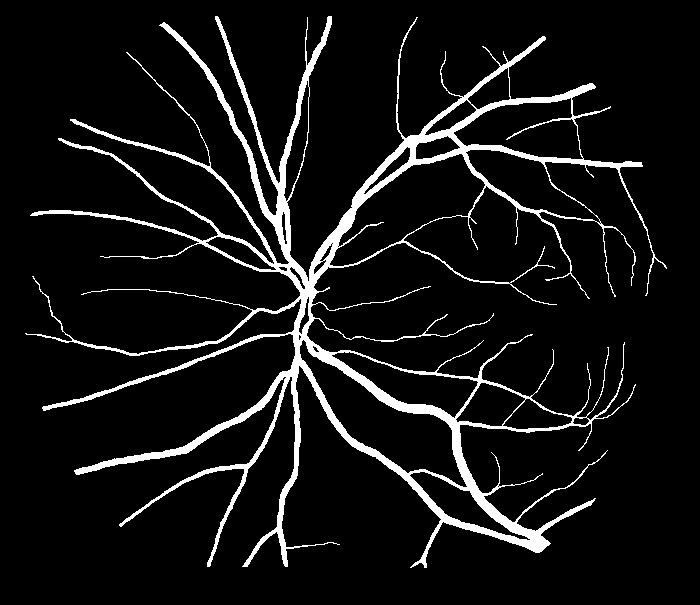}}\hspace{0mm}
\subfloat{\includegraphics[scale=0.12]{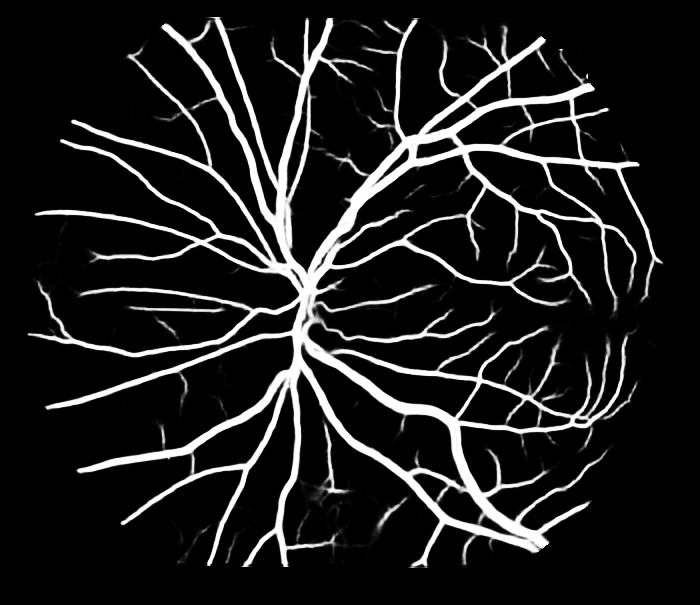}}\hspace{0mm}
\subfloat{\includegraphics[scale=0.12]{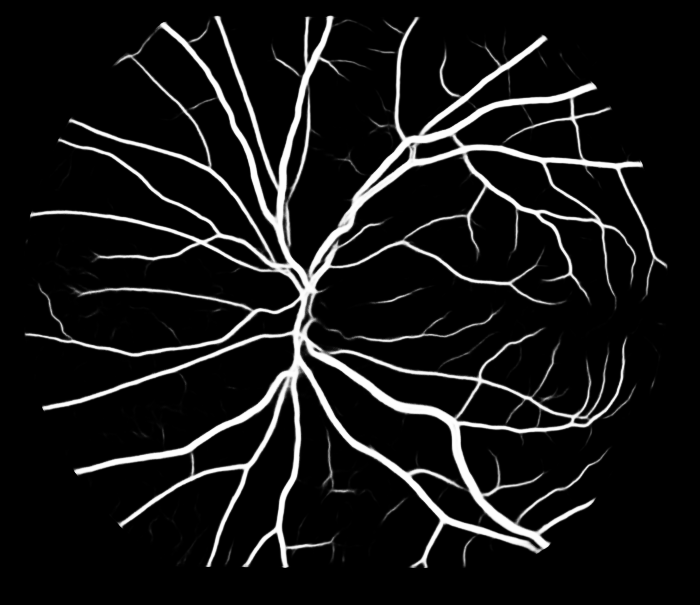}}\hspace{0mm}\\
\hspace{-2mm}
\subfloat{\includegraphics[scale=0.12]{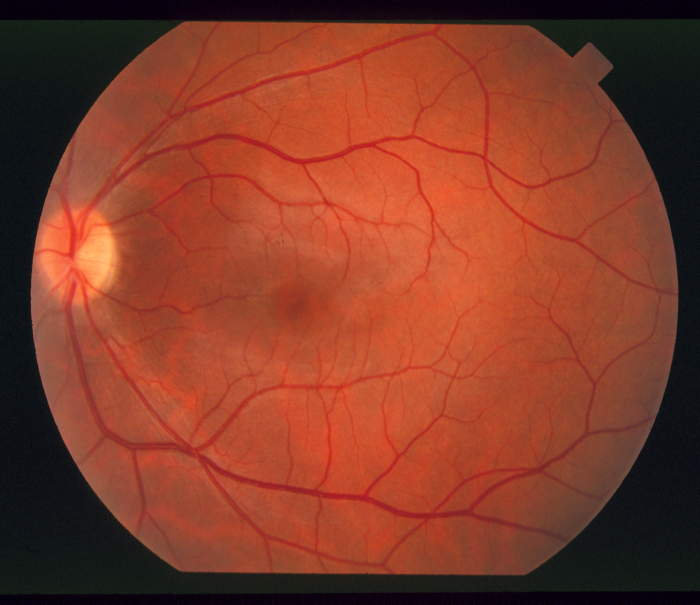}}\hspace{0mm}
\subfloat{\includegraphics[scale=0.12]{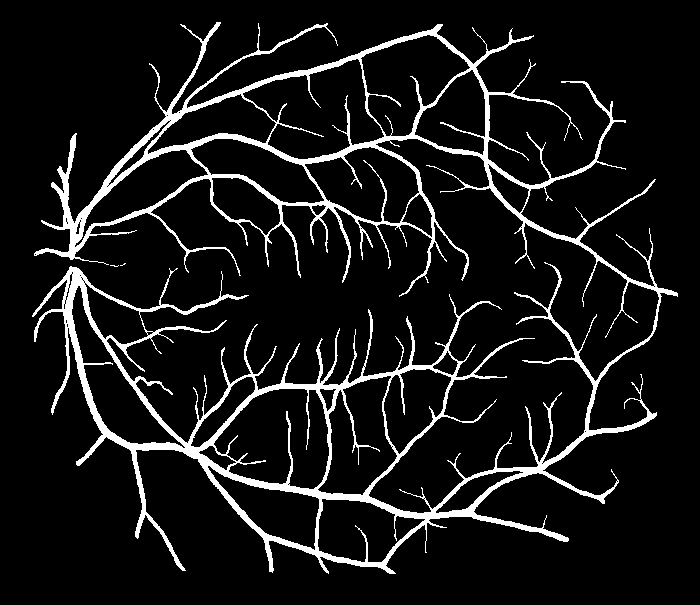}}\hspace{0mm}
\subfloat{\includegraphics[scale=0.12]{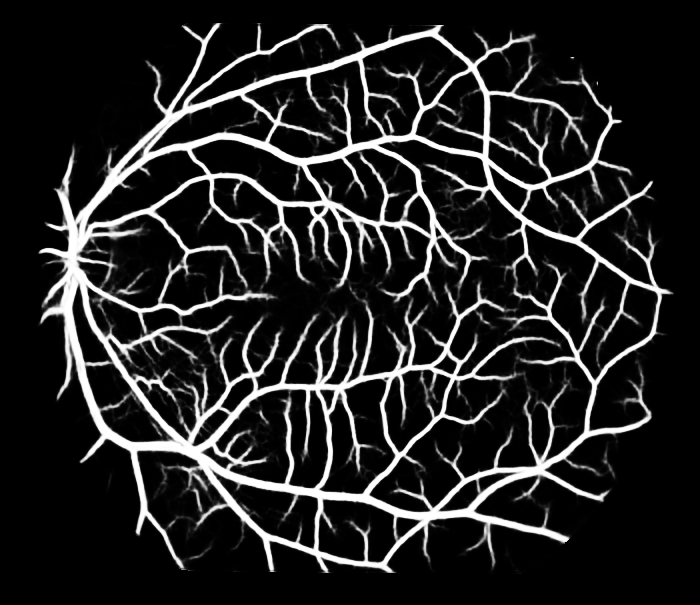}}\hspace{0mm}
\subfloat{\includegraphics[scale=0.12]{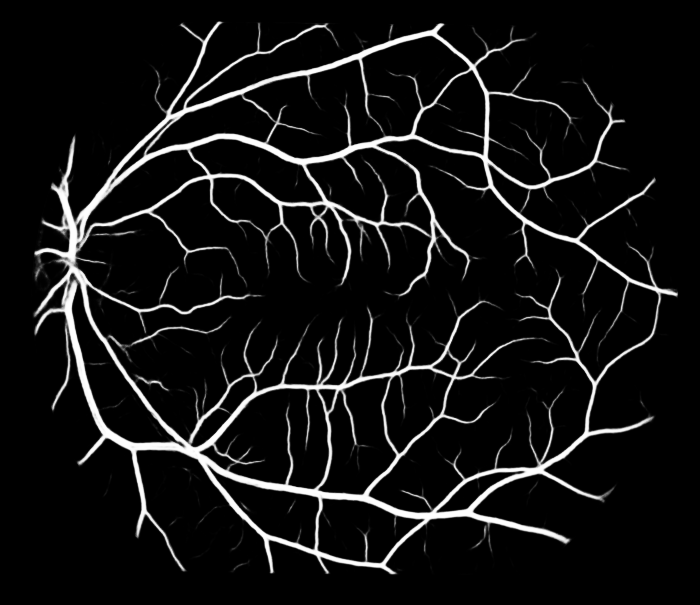}}
\caption{({\bf From left to right}) fundoscopic images, gold standard, probability maps of best existing technique (DRIU~\cite{maninis2016deep}) and probability maps of our method on DRIVE ({\bf top}) and STARE ({\bf bottom}) dataset.}
\label{fig:qualitative_results}
\end{figure}

Fig.~\ref{fig:qualitative_results} illustrates qualitative difference of our method from the best existing method (DRIU). As shown in the figure, our method generates concordant probability maps to the gold standard while DRIU assigns overconfident probability on fine vessels and boundary between vessels and fundus background which may results over-segmentation.

\begin{figure}[!th]
  \centering
\hspace{-2mm}
\subfloat{\includegraphics[scale=0.145]{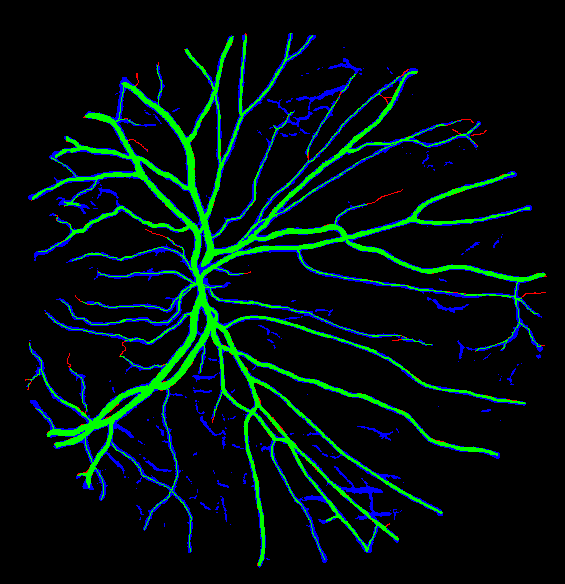}}\hspace{0mm}
\subfloat{\includegraphics[scale=0.145]{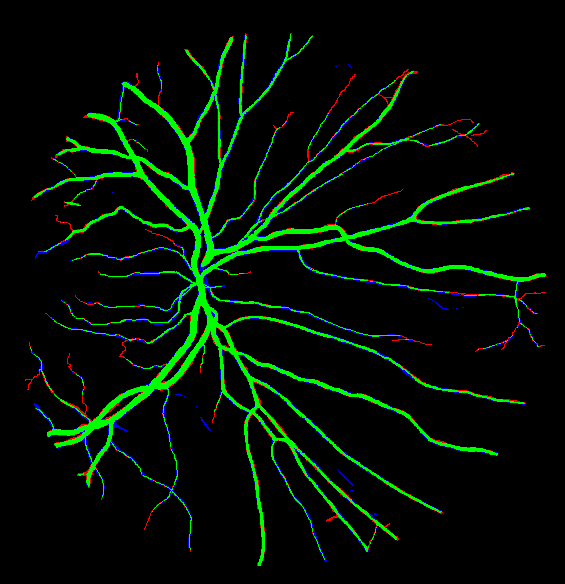}}\hspace{0mm}
\subfloat{\includegraphics[scale=0.145]{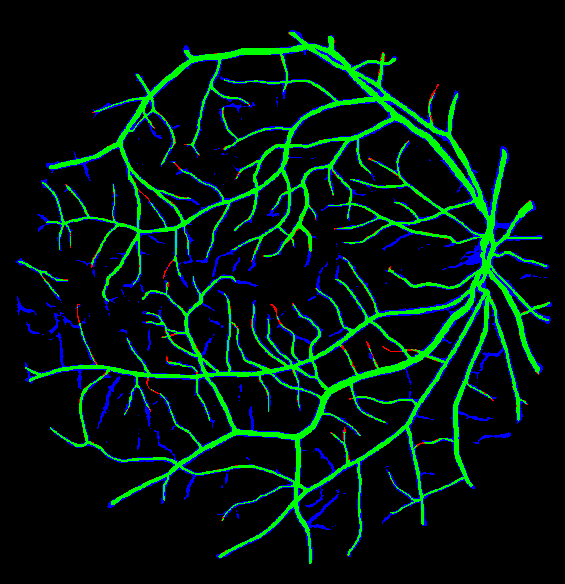}}\hspace{0mm}
\subfloat{\includegraphics[scale=0.145]{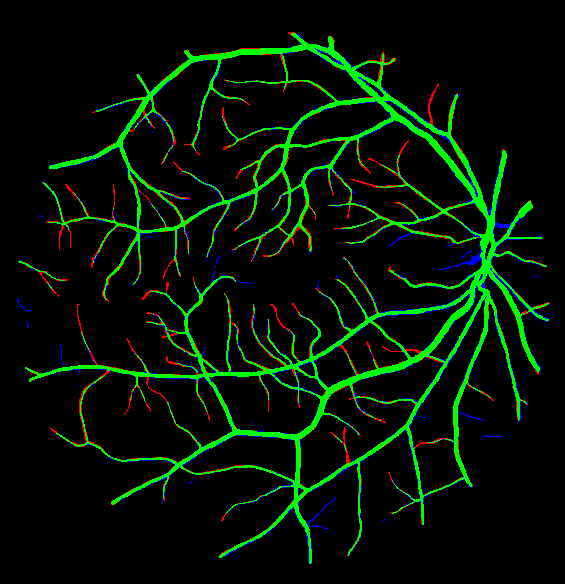}}\hspace{0mm}\\
\noindent\rule{12cm}{1pt}\\
\vspace{3mm}
\hspace{-2mm}
\subfloat{\includegraphics[scale=0.12]{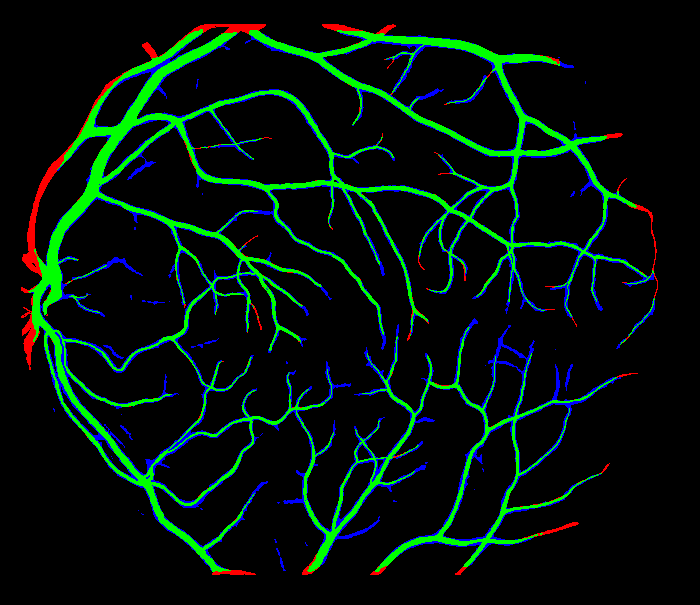}}\hspace{0mm}
\subfloat{\includegraphics[scale=0.12]{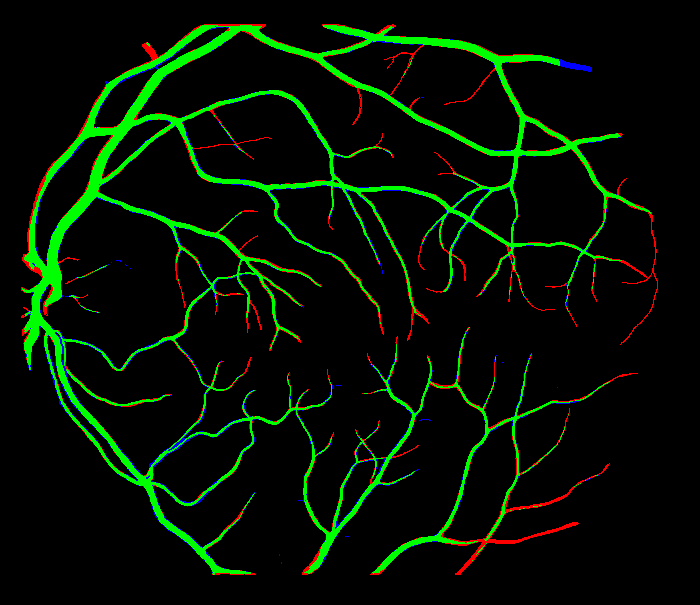}}\hspace{0mm}
\subfloat{\includegraphics[scale=0.12]{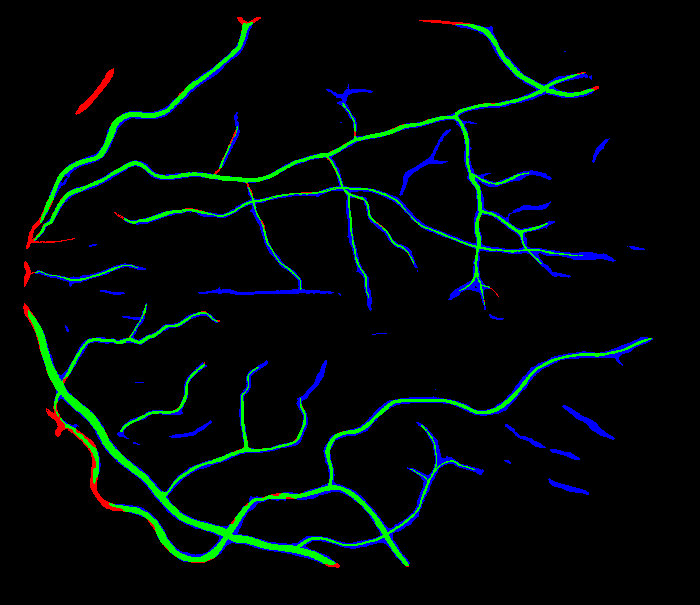}}\hspace{0mm}
\subfloat{\includegraphics[scale=0.12]{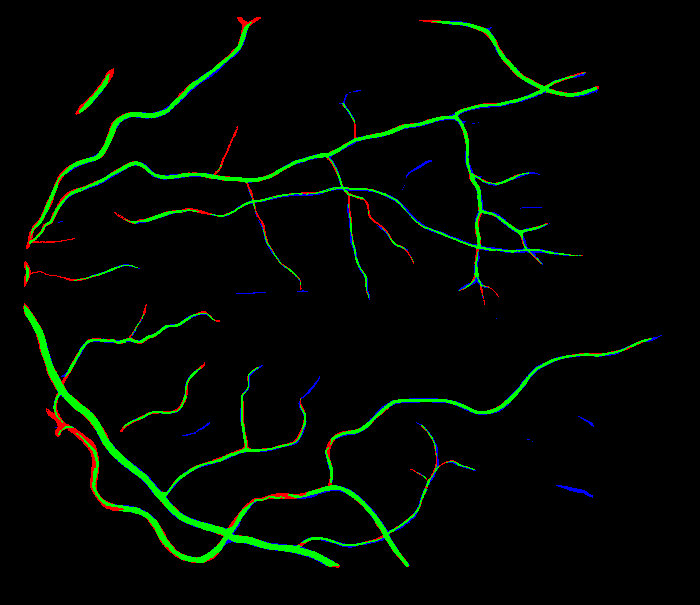}}\hspace{0mm}\\
\caption{Comparison of our method ({\bf 2nd, 4th columns}) with DRIU~\cite{maninis2016deep} ({\bf 1st, 3rd columns}) on DRIVE ({\bf top}) and STARE ({\bf bottom}) dataset. Green marks correct segmentation while blue and red indicate false positive and false negative.}
\label{fig:RGB}
\end{figure}

For further comparison, we converted the probability maps into binary vessel images with Otsu threshold as is done in \cite{fu2016deepvessel}. We can see in Fig~\ref{fig:RGB} that DRIU generally yields more false positives than our method due to the overconfident probability maps. In contrast, our proposed method allows more false negatives around terminal vessels due to its tendency to assign low probability around uncertain regions as human annotators would do.

\section{Conclusion and Discussion}
We introduced GANs framework to retinal vessel segmentation and experimental results suggest that presence of a discriminator can help segment vessels more accurately and clearly. Also, our method outperformed other existing methods in ROC AUC, PR AUC and dice coefficient. Compared to best existing method, our method included less false positives at fine vessels and stroked more clear lines with adequate details like the human annotator. Still, our results fail to detect very thin vessels that span only 1 pixel. We expect that additional prior knowledge on the vessel structures such as connectivity may leverage the performance further. 

\bibliographystyle{splncs03}
\bibliography{retinal_vessel_segmentation}

\begin{thebibliography}{10}
\providecommand{\url}[1]{\texttt{#1}}
\providecommand{\urlprefix}{URL }

\bibitem{becker2013supervised}
Becker, C., Rigamonti, R., Lepetit, V., Fua, P.: Supervised feature learning
  for curvilinear structure segmentation. In: International Conference on
  Medical Image Computing and Computer-Assisted Intervention. pp. 526--533.
  Springer (2013)

\bibitem{fu2016deepvessel}
Fu, H., Xu, Y., Lin, S., Wong, D.W.K., Liu, J.: Deepvessel: Retinal vessel
  segmentation via deep learning and conditional random field. In:
  International Conference on Medical Image Computing and Computer-Assisted
  Intervention. pp. 132--139. Springer (2016)

\bibitem{ganin2014n}
Ganin, Y., Lempitsky, V.: N\^{} 4-fields: Neural network nearest neighbor
  fields for image transforms. In: Asian Conference on Computer Vision. pp.
  536--551. Springer (2014)

\bibitem{goodfellow2014generative}
Goodfellow, I., Pouget-Abadie, J., Mirza, M., Xu, B., Warde-Farley, D., Ozair,
  S., Courville, A., Bengio, Y.: Generative adversarial nets. In: Advances in
  neural information processing systems. pp. 2672--2680 (2014)

\bibitem{he2016deep}
He, K., Zhang, X., Ren, S., Sun, J.: Deep residual learning for image
  recognition. In: Proceedings of the IEEE Conference on Computer Vision and
  Pattern Recognition. pp. 770--778 (2016)

\bibitem{isola2016image}
Isola, P., Zhu, J.Y., Zhou, T., Efros, A.A.: Image-to-image translation with
  conditional adversarial networks. arXiv preprint arXiv:1611.07004  (2016)

\bibitem{maninis2016deep}
Maninis, K.K., Pont-Tuset, J., Arbel{\'a}ez, P., Van~Gool, L.: Deep retinal
  image understanding. In: International Conference on Medical Image Computing
  and Computer-Assisted Intervention. pp. 140--148. Springer (2016)

\bibitem{melinvsvcak2015retinal}
Melin{\v{s}}{\v{c}}ak, M., Prenta{\v{s}}i{\'c}, P., Lon{\v{c}}ari{\'c}, S.:
  Retinal vessel segmentation using deep neural networks. In: VISAPP 2015 (10th
  International Conference on Computer Vision Theory and Applications) (2015)

\bibitem{mirza2014conditional}
Mirza, M., Osindero, S.: Conditional generative adversarial nets. arXiv
  preprint arXiv:1411.1784  (2014)

\bibitem{nguyen2013effective}
Nguyen, U.T., Bhuiyan, A., Park, L.A., Ramamohanarao, K.: An effective retinal
  blood vessel segmentation method using multi-scale line detection. Pattern
  recognition  46(3),  703--715 (2013)

\bibitem{otsu1979threshold}
Otsu, N.: A threshold selection method from gray-level histograms. IEEE
  Transactions on systems, man, and cybernetics  9(1),  62--66 (1979)

\bibitem{radford2015unsupervised}
Radford, A., Metz, L., Chintala, S.: Unsupervised representation learning with
  deep convolutional generative adversarial networks. arXiv preprint
  arXiv:1511.06434  (2015)

\bibitem{ricci2007retinal}
Ricci, E., Perfetti, R.: Retinal blood vessel segmentation using line operators
  and support vector classification. IEEE transactions on medical imaging
  26(10),  1357--1365 (2007)

\bibitem{ronneberger2015u}
Ronneberger, O., Fischer, P., Brox, T.: U-net: Convolutional networks for
  biomedical image segmentation. In: International Conference on Medical Image
  Computing and Computer-Assisted Intervention. pp. 234--241. Springer (2015)

\bibitem{soares2006retinal}
Soares, J.V., Leandro, J.J., Cesar, R.M., Jelinek, H.F., Cree, M.J.: Retinal
  vessel segmentation using the 2-d gabor wavelet and supervised
  classification. IEEE Transactions on medical Imaging  25(9),  1214--1222
  (2006)

\bibitem{xie2015holistically}
Xie, S., Tu, Z.: Holistically-nested edge detection. In: Proceedings of the
  IEEE International Conference on Computer Vision. pp. 1395--1403 (2015)

\bibitem{zhang2010retinal}
Zhang, B., Zhang, L., Zhang, L., Karray, F.: Retinal vessel extraction by
  matched filter with first-order derivative of gaussian. Computers in biology
  and medicine  40(4),  438--445 (2010)

\end{thebibliography}
\end{document}